\documentclass[10pt,twocolumn,letterpaper]{article}

\usepackage[pagenumbers]{cvpr} %

\usepackage{multirow}
\usepackage{makecell}
\usepackage{colortbl}
\usepackage{pifont}  %
\definecolor{RedOrange}{rgb}{1,0.5,0}
\definecolor{BlueGreen}{rgb}{0.0,0.5,0.5}
\definecolor{CadetBlue}{rgb}{0.37,0.62,0.63}
\newcommand{\std}[1]{{\scriptsize\color{gray}$\pm$#1}}             %
\newcommand{\best}[1]{\textbf{#1}}                                   %
\newcommand{\second}[1]{\underline{#1}}                              %
\newcommand{\model}[1]{\texttt{\small #1}}                           %
\newcommand{\red}[1]{\textcolor{RedOrange}{$\uparrow$#1}}            %
\newcommand{\blue}[1]{\textcolor{BlueGreen}{$\downarrow$#1}}         %
\newcommand{\gooddown}[1]{\textcolor{RedOrange}{$\downarrow$#1}}     %
\newcommand{\badup}[1]{\textcolor{BlueGreen}{$\uparrow$#1}}          %
\newcommand{\logo}[1]{\makebox[1.5em][l]{\IfFileExists{logos/#1.pdf}{\raisebox{-0.3\height}{\includegraphics[height=2ex]{logos/#1.pdf}}}{\IfFileExists{logos/#1.png}{\raisebox{-0.3\height}{\includegraphics[height=2ex]{logos/#1.png}}}{}}}}

\definecolor{cvprblue}{rgb}{0.21,0.49,0.74}
\usepackage[pagebackref,breaklinks,colorlinks,allcolors=cvprblue]{hyperref}

\def\confName{CVPR}
\def\confYear{2027}

\title{Efficient Multimodal Agents for Auditing 3D Worlds}

\author{
\textbf{Jingbo Yang}$^{1}$ \quad
\textbf{Ziyan Jiang}$^{1}$ \quad
\textbf{Li An}$^{1}$ \quad
\textbf{Jiabao Ji}$^{1}$ \\
\textbf{Yujian Liu}$^{1}$ \quad
\textbf{Tommi Jaakkola}$^{2}$ \quad
\textbf{Yang Zhang}$^{3}$ \quad
\textbf{Shiyu Chang}$^{1}$ \\
$^{1}$UC Santa Barbara \quad $^{2}$MIT CSAIL \quad $^{3}$MIT-IBM Watson AI Lab
}

\begin{document}
\twocolumn[{%
\renewcommand\twocolumn[1][]{#1}%
\maketitle
\begin{center}
  \captionsetup{type=figure}
  \includegraphics[width=\textwidth]{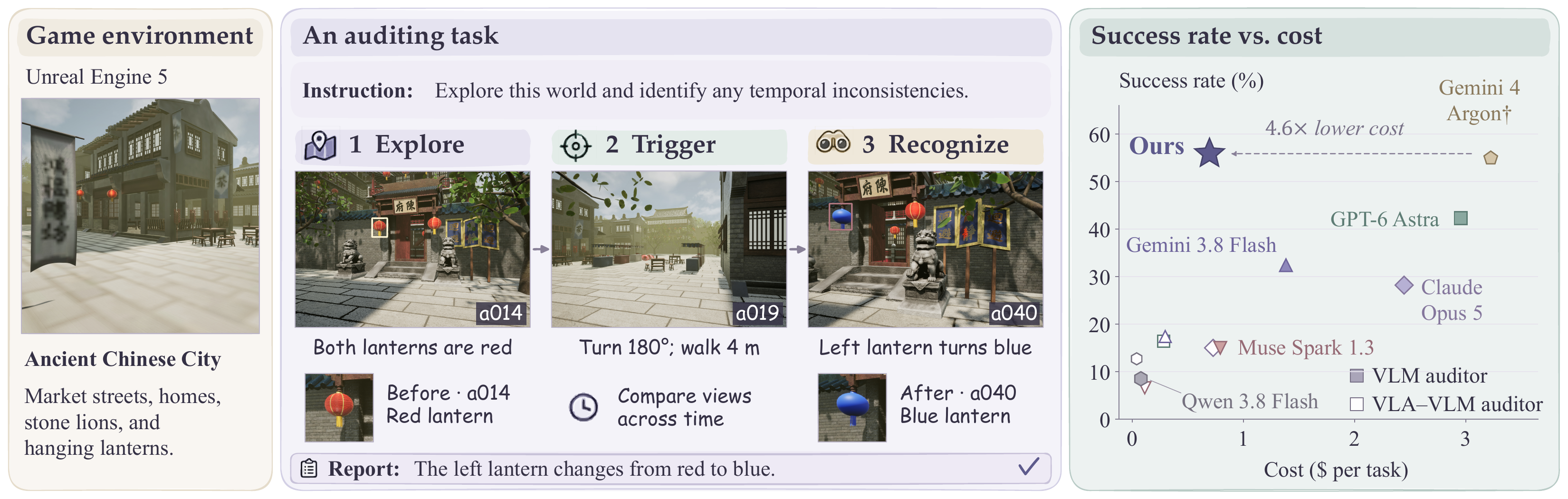}
  \captionof{figure}{\textbf{Auditing a 3D world.} In an Unreal Engine~5 world, the auditor has to bring the lanterns into view
  (explore), turn away and come back (trigger), and notice that the left lantern has turned blue (recognize). The plot shows success
  rate against cost per task on WorldAuditBench (213 tasks, list prices), with one marker per VLM, filled for VLM auditors and hollow
  for VLA--VLM auditors.}
  \label{fig:cost}
\end{center}%
\vspace{0.6em}
}]
\begin{abstract}
Multimodal agents can now build interactive 3D worlds and games, yet these worlds, like those built by people, often contain glitches such as floating objects, walls that can be walked through, or objects that change when no one is looking. An automatic auditor could check such worlds at scale and provide rewards for agents that build them, but only if it is both accurate and cheap. Auditing requires three abilities, namely navigating to the objects that may contain a glitch, triggering glitches that appear only after a specific behavior, and recognizing what is wrong. Current auditors are either accurate but slow and expensive or cheap but unreliable. We study how much of the audit small models can take on. A 4B vision language model fine-tuned end to end on verified trajectories of a frontier teacher guided by Bayesian search learns to navigate and trigger, and six such explorers together match a frontier model on both, but recognition stays far behind. We therefore build an auditor in which a swarm of small explorers navigates and triggers, and a frontier model reviews what they record and report. On WorldAuditBench, our auditor succeeds on 55.9\% of 213 tasks in Unreal Engine~5 and Three.js worlds, 13.6 points above the strongest auditor reported, at less than a quarter of its cost. It matches the frontier reviewer exploring on its own at 4.6$\times$ lower cost and about half the latency. Without further training, it also answers embodied questions about real-world scans, matching the frontier reviewer exploring on its own and surpassing four other frontier models by 6.4 to 16.6 points. Code is available at \url{https://github.com/UCSB-NLP-Chang/WorldAuditor}.
\end{abstract}
    
\section{Introduction}
\label{sec:intro}

With advances in the multimodal coding capabilities of vision-language models (VLMs), AI agents can now synthesize interactive 3D worlds and games directly from simple user instructions~\cite{yang2024holodeck,hu2024scenecraft,sun20253d,bruce2024genie}. Much like human-crafted environments, however, these generated worlds often exhibit glitches: a chair floats midair, a wall lacks collision boundaries and allows player pass-through, or a lamp spontaneously changes color unobserved. In conventional game studios, quality assurance testers identify such anomalies by manually playing through every scene—a process that is both labor-intensive and costly~\cite{politowski2021survey}. World-generating agents exacerbate this bottleneck by producing environments far faster than human evaluators can inspect them. Consequently, recent efforts have focused on developing VLM-based auditing agents that can autonomously navigate 3D environments and detect glitches~\cite{jiang2026worldauditbench}. Such auditors can steer the iterative debugging of synthetic worlds~\cite{madaan2023self,zheng2023judging} or act as automated reward models for training world-generation agents~\cite{rocamonde2024vision}.

Comprehensive 3D world auditing demands three core competencies: (1)~\textit{navigation}, wherein the auditor navigates the scene to discover anomalies, which involves identifying highly suspicious areas, formulating an inspection route, and executing traversal using primitive actions;
(2)~\textit{triggering}, as many glitches manifest only after specific interactions (e.g., walking into an uncollided mesh or looking away from an object and back), requiring the auditor to understand how to elicit each fault type; and (3)~\textit{recognition}, where the auditor must visually identify and verify the anomaly once it enters the field of view. 

As 3D world auditing requires both high accuracy and computational efficiency, a viable strategy is to decompose the audit along the three competencies above and assign each to models of appropriate sizes. Existing work~\citep{jiang2026worldauditbench} compares two primary paradigms. In the first paradigm, a single frontier VLM executes all three competencies end to end. Although accurate, this approach is computationally expensive. On WorldAuditBench~\cite{jiang2026worldauditbench}, a single audit by a frontier VLM exceeds ten minutes and costs approximately \$3.00. In the second paradigm, an inexpensive vision-language-action (VLA) model~\cite{yue2026scaling} explores the environment and logs visual observations, which a frontier VLM subsequently inspects offline. While cost-effective, current VLA models lack the reasoning needed to prioritize inspection targets or trigger latent glitches, leading to frequent misses and false alarms. Consequently, the strongest such auditor achieves a success rate of only 17.4\%.

In short, neither existing paradigm achieves a desirable trade-off between accuracy and efficiency. Although the second paradigm moves in a promising direction by offloading exploration to a lightweight model, the chosen VLA policy lacks the reasoning capabilities required across the three auditing competencies. This observation motivates two core \textbf{research questions:} (1)~Can lightweight models, when specifically optimized for auditing, enable a better trade-off between accuracy and efficiency in 3D world auditing? (2)~How should auditing competencies be divided and coordinated across small and frontier models to achieve an optimal trade-off?

To answer these questions, we conduct two studies. To address the first question, we fine-tune a small VLM end to end on the full auditing pipeline, covering navigation, triggering, and reporting, and evaluate its proficiency across all three competencies. We construct training data in worlds with injected anomalies, where a frontier teacher explores using a strategy grounded in Bayesian search theory~\cite{koopman1956theory1,stone1976theory} and the game engine verifies every retained trajectory. To address the second question, we evaluate multiple ways to partition the audit between small and frontier models, comparing a frontier auditor alone, a small auditor alone, an ensemble of small auditors pooling their reports, and a swarm of small explorers with a frontier reviewer.

\begin{figure*}[t]
  \centering
  \includegraphics[width=\textwidth]{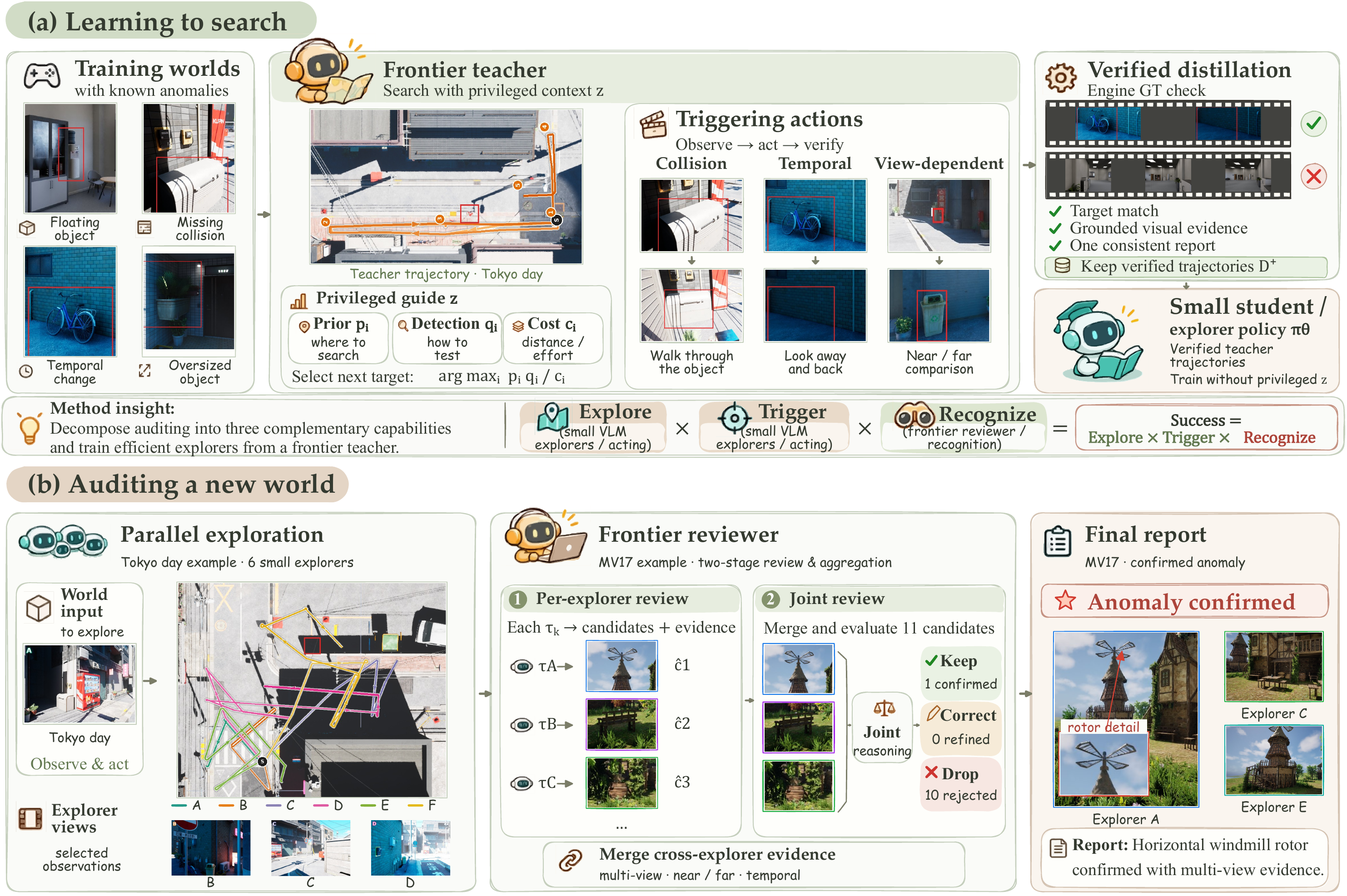}
  \caption{\textbf{Overview.} (a) Learning to search (\cref{sec:method:training}). A frontier teacher audits training worlds with
  injected anomalies and receives a privileged context $z$ that states Bayesian search. Trajectories that pass the check against the
  ground truth of the engine train the small explorer policy $\pi_\theta$, which never sees $z$. (b) Auditing a new world
  (\cref{sec:method:swarm,sec:method:review}). Six small explorers explore in parallel from the start pose. A frontier reviewer reviews each trajectory,
  then all candidates jointly, and reports the anomalies that the explorers' frames support.}
  \label{fig:method}
\end{figure*}

Surprisingly, our most effective design never requires the frontier model to act in the environment. Instead, a swarm of lightweight explorers handles navigation and triggering in parallel, while the frontier model exclusively reviews their recorded trajectories and preliminary reports (\cref{fig:method}). After fine-tuning, a single small explorer still recognizes few glitches on its own, but six parallel explorers collectively bring more glitches into view and trigger a higher fraction of them than Gemini~4 Argon exploring alone. Consequently, reviewing their pooled recordings uncovers as many glitches as a standalone frontier agent. Across 213 tasks in Unreal Engine~5 and Three.js environments on WorldAuditBench, our auditor solves 55.9\% of tasks, outperforming the strongest reported baseline by 13.6 points at less than a quarter of its cost (\cref{fig:cost}). It matches standalone Gemini~4 Argon (54.9\%) at 4.6$\times$ lower cost and roughly half the latency. Moreover, varying the number of explorers controls the trade-off between accuracy and cost, with a single explorer already solving 35.8\% of tasks at \$0.19 per task. Without any task-specific training, the same system also transfers zero-shot to embodied question answering on real-world 3D scans, matching Gemini~4 Argon on A-EQA~\cite{majumdar2024openeqa} and outperforming four other frontier models by 6.4 to 16.6 points.

\section{Related Work}
\label{sec:related}

\paragraph{Game testing with AI.}
Automated game testing has long used agents that explore a game. Agents trained with reinforcement learning or search cover maps, reach unexpected states, and expose exploits~\cite{zheng2019wuji,bergdahl2020augmenting,gordillo2021improving,sestini2022automated,lu2022go,liu2022inspector}, but they are trained for each game, read its internal state, and do not describe what is wrong. Vision language models have since been studied as glitch detectors. Most benchmarks give the model screenshots or gameplay videos and ask it to find or describe a glitch~\cite{taesiri2022clip,taesiri2024glitchbench,cao2024physgame,taesiri2026videogameqa,zheng2026open,yu2026tempglitch,lu2026far}, and fine-tuning on game images helps it answer such questions~\cite{taesiri2025videogamebunny}. These benchmarks test recognition alone, since the model never chooses where to look or what to try. Other work lets LLM agents act in a game. GBQA~\cite{jiang2026gbqa} tests web games of six genres through text observations and API calls, TITAN~\cite{wang2025leveraging} tests online role-playing games through structured game state, and GUI agents play generated 2D browser games to check their behavior~\cite{huang2026gui}. In 3D, an explorer that covers a level can collect frames in which VLMs look for clipping~\cite{celemin2026evaluating}, and World of Bugs~\cite{wilkins2022world} provides 3D scenes with bugs for learned detectors. To our knowledge, WorldAuditBench~\cite{jiang2026worldauditbench} is the only benchmark in which a multimodal agent explores interactive 3D worlds, triggers their glitches, and reports them with evidence, so we evaluate on it.

\paragraph{Building 3D worlds and games with AI.}
Language and vision language models now build 3D scenes and whole worlds. They plan layouts~\cite{yang2024holodeck,sun2025layoutvlm}, write code for a renderer or a game engine~\cite{sun20253d,hu2024scenecraft,duan2025latticeworld,kang2026simworld}, and critique and repair their own scenes~\cite{yang2026sceneweaver,xia2026sage,ling2026scenethesis}. Coding agents build playable games in Pygame, Three.js, Godot, and Unreal Engine~\cite{zhang2026v,bakshi2026worldbench,lu2606worldcoder,luo2026gamecraft,yin2026autoue}, and world models generate interactive worlds frame by frame~\cite{bruce2024genie,valevski2025diffusion,parker2025genie,he2025matrix}. Their evaluations check whether the requested content is there, from the code, runtime state, screenshots, or short tours of the world~\cite{zhang2026artifactsbench,lu2606worldcoder,luo2026gamecraft,bakshi2026worldbench}. A wall that lets the player through or an object that changes while unobserved passes these checks, so what these systems build still needs an auditor.

\section{Method}
\label{sec:method}

We first define the auditing task and relate its three core abilities to auditor success (\cref{sec:method:problem}). We then analyze how to divide these abilities between small and frontier models (\cref{sec:method:division}). Because the optimal division depends on small-model capabilities, we train a lightweight model end to end on the complete audit (\cref{sec:method:training}) and design our auditor accordingly (\cref{sec:method:swarm,sec:method:review}).

\subsection{Auditing a 3D World}
\label{sec:method:problem}

An auditing task consists of a 3D world and a task instruction. The world contains a set $X^*$ of anomalies, elements whose geometry, physics, appearance, or temporal behavior violate physical plausibility or scene semantics. $X^*$ may contain several anomalies or none, and their types are unknown in advance. The instruction describes the scene and may provide hints, such as anomaly types to look for. At step $t$, the auditor receives an observation $o_t$, comprising a first-person view and the text result of its previous tool call, and issues an action $a_t$. The tool set includes primitive navigation actions (\texttt{move}, \texttt{turn}, \texttt{look}, \texttt{wait}, \texttt{interact}), memory retrieval tools (\texttt{inspect}, \texttt{history}), and bug-reporting actions. An episode allows at most $B$ environment actions and concludes with a report $r$, a set of findings that each describe an anomaly and cite supporting frames as evidence. An anomaly $x \in X^*$ is found ($S_x{=}1$) when a finding identifies it and cites frames showing it.

Finding an anomaly $x$ requires three sequential abilities. First, bug-seeking navigation must bring $x$ into view (event $E$). Second, triggering must perform the action that exposes $x$ (event $T$), such as walking into an object without collision or looking away from an object and back when it changes while unobserved. For anomalies that appear passively, $T$ coincides with $E$. Third, recognition must identify $x$ in the report (event $R$). Hence:
\begin{equation}
  P(S_x) = P(E)\, P(T \mid E)\, P(R \mid T).
  \label{eq:decomp}
\end{equation}

\subsection{Dividing the Audit}
\label{sec:method:division}

The factors in \cref{eq:decomp} need not come from the same model. We refer to any model that acts in the world, and thus handles navigation and triggering, as an explorer. A frontier VLM is a capable explorer but incurs high cost and latency, as it selects every action while repeatedly reprocessing an expanding image context (\cref{sec:exp:efficiency}). In contrast, lightweight VLMs are inexpensive and can be deployed in parallel across multiple instances. We therefore investigate three paradigms that delegate an increasing share of the audit to small models. \ding{182}~A single small model conducts the audit end to end, handling navigation, triggering, and reporting on its own (\cref{sec:method:training}). \ding{183}~A swarm of small explorers searches the world in parallel and pools their individual reports (\cref{sec:method:swarm}). \ding{184}~A frontier model reviews the explorers' recorded trajectories offline and identifies the anomalies they capture, without acting in the world (\cref{sec:method:review}).

\subsection{Training a Small Auditor}
\label{sec:method:training}

We fine-tune a small VLM on trajectories from a frontier teacher auditing worlds with injected anomalies, keeping only trajectories verified by engine ground truth (\cref{fig:method}a).

\paragraph{Worlds with known anomalies.}
\label{sec:method:worlds}
In Three.js, a procedural pipeline constructs seven scene types, such as houses, offices, schools, and streets. An LLM plans each layout, the generator places functional object groups on wall and surface anchors, and automated physics checks and a VLM filter prune misplaced objects. We generate 55 such worlds for training. In Unreal Engine~5 (UE5), we use three scenes across two commercial environments: an office building and a street district under daytime and nighttime conditions (\cref{sec:app:worlds}). Our anomaly taxonomy follows established game bug classifications~\cite{lewis2010went,truelove2021we,butt2023deriving}. We implement injectors using Claude Code~\cite{anthropic2025claudecode}, covering 44 anomaly types across five categories: geometry, collision, appearance, temporal, and semantic (\cref{sec:app:types}). For every frame, the engine logs the visible target area and whether triggering events occurred (e.g., player penetration into collision-free meshes).

\paragraph{A teacher guided by Bayesian search.}
\label{sec:method:teacher}
The teacher, Gemini~4 Argon, audits training worlds using the same tools and instructions as the test-time auditor, along with a privileged search context $z$. Bayesian search theory~\cite{koopman1956theory1,koopman1956theory2,koopman1957theory3,stone1976theory} models a target hidden in one of $n$ cells with prior probabilities $p_i$, where inspecting cell $i$ costs $c_i$ and detects the target with probability $q_i$. For a stationary target, the expected search cost is minimized by iteratively checking the cell with the highest ratio~\cite{matula1964periodic,stone1976theory}:
\begin{equation}
  i^* = \arg\max_i \; \frac{p_i\, q_i}{c_i},
  \label{eq:search}
\end{equation}
updating beliefs via Bayes' rule after each unsuccessful check. The context $z$ formulates this strategy in natural language, treating candidate objects as cells and targeted checks as search attempts. Anomaly cards specify the prior $p$ based on typical defect locations and suggest the check with highest detection probability $q$ (e.g., walking into a fence to test collision). Walking distance from the current position serves as cost $c$. The teacher estimates these values from visual input, inspects candidates following \cref{eq:search}, and skips verified objects. Before each action, it articulates what it sees and what the action tests (\cref{sec:app:teacher}). The student never receives $z$ and must learn search strategies purely from observations, following learning with privileged information~\cite{vapnik2009new,chen2020learning} and context distillation~\cite{askell2021general,snell2022learning}. Trajectories whose text reveals $z$ are discarded.

\paragraph{Verified distillation.}
\label{sec:method:distill}
Engine ground truth governs trajectory acceptance. A trajectory is accepted if its report matches injected anomalies bijectively: every finding identifies an injected anomaly with verified visual evidence, and anomaly-free worlds produce empty reports. Static anomalies need a frame with sufficient target visibility, temporal anomalies need frames before and after the change, and interactive anomalies need a logged trigger (\cref{sec:app:verify}). Only 31\% of the teacher's Three.js trajectories pass, from 98\% for changes of daylight to 11\% for missing collision, and the engine's logs also catch claimed tests that never happened, such as walking through an object it only passed.

Procedural worlds can also harbor uninjected artifacts. If the teacher reports such a bug, it resembles a false alarm; if ignored, the student learns to overlook real defects. Therefore, a VLM inspects close-up renders of non-injected objects flagged by the teacher, and verified artifacts are repaired before generating training trajectories. This ensures the training corpus contains only clean demonstrations.

Denoting accepted trajectories as $S{=}1$, filtering samples from the teacher conditioned on success, $p_T(\tau \mid S{=}1, z)$, following rejection sampling~\cite{anthony2017thinking,zelikman2022star,gulcehre2023reinforced}. Let $h_t$ be the instruction and history up to turn $t$ without $z$, and $y_t$ the teacher's response. The student $\pi_\theta$ minimizes:
\begin{equation}
  \mathcal{L}(\theta) = -\,\mathbb{E}_{\tau \sim p_T(\cdot \mid S=1,\, z)} \sum_t \log \pi_\theta(y_t \mid h_t),
  \label{eq:sft}
\end{equation}
learning navigation, triggering, and reporting end to end. The final dataset includes 2,599 accepted trajectories (1,798 in Three.js, 801 in UE5) across all 44 types (\cref{sec:exp:setup}). After training, the small model brings 72.3\% of the anomalies into view and triggers 65.7\%, close to Gemini~4 Argon (88.7\% and 85.0\%), but recognizes only 19.0\% of those it triggers, against 61.3\% (\cref{sec:exp:analysis}), so paradigm \ding{182} alone solves only 15.7\% of the tasks.

\subsection{Swarm of Explorers}
\label{sec:method:swarm}
\label{sec:method:auditor}

Because a small explorer is cheap, several can search the same world in parallel, and six of them together reach and trigger more anomalies than the frontier model (\cref{sec:exp:analysis}). All $K$ explorers share the trained policy $\pi_\theta$. They initialize from the task starting pose, sample actions independently, and operate within the same action budget and tool space as a single auditor. Explorer $k$ returns trajectory $\tau_k$, containing captured frames, action logs with coordinates and collision events, and preliminary reports, and in paradigm \ding{183} the auditor outputs the union of these reports. If one explorer triggers an anomaly with probability $\rho$, then $K$ independent explorers trigger it with probability
\begin{equation}
  P_K(T) = 1 - (1 - \rho)^K.
  \label{eq:coverage}
\end{equation}

\subsection{Frontier Review}
\label{sec:method:review}

To recognize what the explorers miss, a frontier model inspects their trajectories offline without environment interaction. Verification follows a two-stage process. First, the reviewer evaluates each trajectory and its preliminary reports individually, extracting candidate anomalies with corresponding keyframes. Next, it jointly reviews the aggregated candidate pool across all explorers, merges duplicate findings, and filters false positives based on visual evidence. The joint review sees a fixed budget of frames, mostly the evidence around reported candidates, with the rest covering the explored area. Surviving candidates form the final report. Inspecting individual trajectories first keeps each model call concise, while joint review enables cross-validation across multiple explorer viewpoints (\cref{sec:exp:setup}). Because the reviewer processes each trajectory once, its cost scales linearly with $K$ and trajectory length. \cref{sec:exp:ablation} compares the three paradigms.

\section{Experiments}
\label{sec:exp}

\begin{table*}[t]
  \centering
  \setlength{\tabcolsep}{5.4pt}
  \renewcommand{\arraystretch}{1.05}
  \caption{\textbf{Main results on WorldAuditBench.} Success rate (\%) by anomaly family and engine, overall success rate (SR), and
  cost per task at list prices. Rows with $^\dagger$ are run by us, and the other rows are copied from
  WorldAuditBench~\cite{jiang2026worldauditbench}. $\Delta$ compares ours with each row (\textcolor{RedOrange}{orange}: ours better,
  \textcolor{BlueGreen}{teal}: ours worse). \textbf{Bold} and \underline{underline} mark the best and second best.}
  \label{tab:main}
  \footnotesize
  \renewcommand{\model}[1]{\texttt{#1}}%
  \begin{tabular}{cl|ccccc|cc|cc|cc}
    \toprule
    \rowcolor{CadetBlue!15}
    \cellcolor{white} & \cellcolor{white}
    & \multicolumn{5}{c|}{\textbf{Anomaly family}}
    & \multicolumn{2}{c|}{\textbf{Engine}}
    & \multicolumn{2}{c|}{\textbf{Overall}}
    & \multicolumn{2}{c}{\textbf{Cost} (\$/task)} \\
    \cmidrule(lr){3-7} \cmidrule(lr){8-9} \cmidrule(lr){10-11} \cmidrule(lr){12-13}
    \cellcolor{white} & \cellcolor{white} \multirow{-2.25}{*}{\textbf{Method}}
    & SP & IP & SpC & TC & SeC & UE5 & Three.js & SR$\uparrow$ & $\Delta$ & \$$\downarrow$ & $\Delta$ \\
    \midrule
    \multirow{6}{*}{\makecell{\textsc{VLM}\\\textsc{Auditors}}}
    & \logo{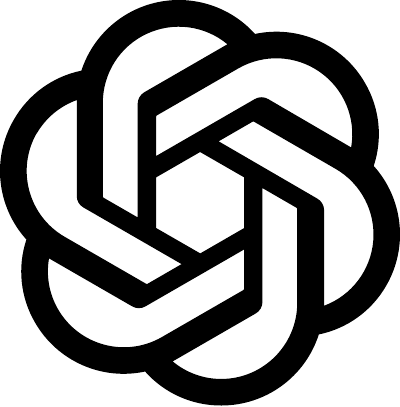}\model{GPT-6 Astra} & 59.3 & 29.3 & 45.1 & 12.5 & \best{68.2} & 47.6 & 34.5 & 42.3 & \red{13.6} & 2.955 & \gooddown{2.26} \\
    & \logo{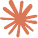}\model{Claude Opus 5} & 35.6 & 19.5 & 29.4 & 12.5 & 50.0 & 27.8 & 28.7 & 28.2 & \red{27.7} & 2.445 & \gooddown{1.75} \\
    & \logo{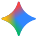}\model{Gemini 3.8 Flash} & 49.2 & 22.0 & 29.4 & 12.5 & 50.0 & 33.3 & 31.0 & 32.4 & \red{23.5} & 1.382 & \gooddown{0.69} \\
    & \logo{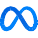}\model{Muse Spark 1.3} & 22.0 & 12.2 & 13.7 & \phantom{0}5.0 & 22.7 & 11.1 & 20.7 & 15.0 & \red{40.9} & 0.791 & \gooddown{0.10} \\
    & \logo{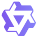}\model{Qwen 3.8 Flash} & 11.9 & \phantom{0}7.3 & \phantom{0}3.9 & \phantom{0}2.5 & 22.7 & \phantom{0}8.7 & \phantom{0}8.0 & \phantom{0}8.5 & \red{47.4} & 0.077 & \badup{0.62} \\
    & \logo{gemini}\model{Gemini 4 Argon}$^\dagger$ & \second{74.6} & \second{43.9} & \best{64.7} & \second{20.0} & \second{63.6} & \second{57.9} & \second{50.6} & \second{54.9} & \red{1.0} & 3.225 & \gooddown{2.53} \\
    \midrule
    \multirow{5}{*}{\makecell{\textsc{VLA--VLM}\\\textsc{Auditors}}}
    & \logo{openai}\model{GPT-6 Astra} & 20.3 & 17.1 & 25.5 & \phantom{0}0.0 & 13.6 & \phantom{0}5.6 & 32.2 & 16.4 & \red{39.5} & 0.281 & \badup{0.41} \\
    & \logo{claude}\model{Claude Opus 5} & 22.0 & 22.0 & 11.8 & \phantom{0}2.5 & 13.6 & \phantom{0}2.4 & 33.3 & 15.0 & \red{40.9} & 0.727 & \gooddown{0.03} \\
    & \logo{gemini}\model{Gemini 3.8 Flash} & 20.3 & 24.4 & 17.6 & \phantom{0}5.0 & 18.2 & \phantom{0}5.6 & 34.5 & 17.4 & \red{38.5} & 0.296 & \badup{0.40} \\
    & \logo{meta}\model{Muse Spark 1.3} & \phantom{0}5.1 & \phantom{0}9.8 & \phantom{0}9.8 & \phantom{0}0.0 & \phantom{0}9.1 & \phantom{0}2.4 & 12.6 & \phantom{0}6.6 & \red{49.3} & 0.112 & \badup{0.58} \\
    & \logo{qwen}\model{Qwen 3.8 Flash} & 18.6 & 22.0 & 11.8 & \phantom{0}0.0 & \phantom{0}4.5 & \phantom{0}5.6 & 23.0 & 12.7 & \red{43.2} & 0.039 & \badup{0.65} \\
    \midrule
    \rowcolor{gray!10}
    \multicolumn{2}{c|}{\textsc{Ours}} & \best{81.4} & \best{46.3} & \second{56.9} & \best{22.5} & \second{63.6} & \best{58.7} & \best{51.7} & \best{55.9} & -- & 0.694 & -- \\
    \bottomrule
  \end{tabular}
\end{table*}

\subsection{Experimental Setup}
\label{sec:exp:setup}

\paragraph{Benchmark.}
WorldAuditBench contains 213 auditing tasks in 13 environments, 126 built with Unreal Engine~5 (UE5) and 87 with Three.js.
Each task hides one anomaly from one of five families:
static physics (SP, 59 tasks), interactive physics (IP, 41), spatial consistency (SpC, 51),
temporal consistency (TC, 40), and semantic consistency (SeC, 22).
We follow the benchmark protocol: every auditor receives the scene description, the anomaly-type hint, and the in-context examples,
may take at most 40 environment actions, and reports anomalies with supporting screenshots.
No benchmark environment appears in our training data. Our UE5 maps come from two other commercial environments
and our Three.js worlds from our own generator.

\paragraph{Metrics.}
As in WorldAuditBench, GPT-6 Astra judges each report against the task rubric, and a task succeeds if any reported finding matches the hidden anomaly.
We report the success rate (SR) by anomaly family, by engine, and overall.
Cost is the inference cost per task in US dollars at list prices, without prompt-caching discounts; judge calls are excluded.
Our explorers run locally; we price their tokens conservatively at the OpenRouter list price of Qwen3.5-9B (\cref{sec:app:cost}).
Latency is analyzed in \cref{sec:exp:efficiency}.

\paragraph{Baselines.}
We copy the results of both auditing paradigms in WorldAuditBench.
\emph{VLM auditors} (GPT-6 Astra, Claude Opus~5, Gemini~3.8 Flash, Muse Spark~1.3, and Qwen~3.8 Flash) choose every action themselves,
each running in its vendor's agent harness.
\emph{VLA--VLM auditors} explore with the Open-P2P 1.2B VLA model~\cite{yue2026scaling} and let each VLM analyze the recorded trajectory.
We additionally run two VLM auditors under the same protocol (marked $^\dagger$):
Gemini~4 Argon, the model our method uses for review, and the untrained Qwen3.5-4B, the base model of our explorers.

\paragraph{Implementation details.}
Explorers are Qwen3.5-4B fine-tuned on the data of \cref{sec:method:distill} for three epochs (learning rate $2{\times}10^{-5}$, batch size 16,
contexts up to 44,000 tokens, which leaves out the longest 5\% of the trajectories) and served with SGLang~\cite{zheng2024sglang}
(temperature 0.7, top-$p$ 0.8, top-$k$ 20, thinking disabled, 640-pixel-wide frames).
Unless stated otherwise, $K{=}6$ explorers start from the benchmark start pose and explore independently with the same 40-action budget.
Gemini~4 Argon first reviews each explorer's trajectory (the final view of every action, the frames it cites, and the action log, at medium reasoning effort and low media resolution)
and proposes up to three candidate anomalies.
A final review with high reasoning effort sees all candidates, the action logs, and 96 frames; it keeps, corrects, or drops each candidate
and reports at most three anomalies.
Each configuration runs once unless a standard deviation is given; repeating the review three times on the same
trajectories gives 56.3\,$\pm$\,0.8\% overall (\cref{sec:app:variance}).

\subsection{Main Results}
\label{sec:exp:main}

\paragraph{More successful and cheaper than WorldAuditBench auditors.}
With six explorers, our method succeeds on 55.9\% of tasks (\cref{tab:main}), 13.6 points above the strongest auditor in WorldAuditBench
(GPT-6 Astra, 42.3\%), at less than a quarter of its cost (\$0.69 \vs \$2.96 per task).
It has the highest success rate of all auditors overall and on each engine (58.7\% on UE5, 51.7\% on Three.js),
and it is best or second best in seven of the ten engine--family columns.
Our method also does well where a glitch has to be triggered. On interactive physics, it solves 46.3\% of the tasks, the best of all
auditors, against 29.3\% for GPT-6 Astra and 43.9\% for Gemini~4 Argon exploring alone.
VLA--VLM auditors cost \$0.04--0.73 per task but succeed on at most 17.4\% of tasks.
Even with a single explorer, our method reaches 35.8\% at \$0.19 per task (\cref{fig:cost}), above every WorldAuditBench auditor except GPT-6 Astra.

\paragraph{Comparison with Gemini~4 Argon.}
Our reviewer model is itself the strongest baseline: exploring alone, Gemini~4 Argon succeeds on 54.9\% of tasks.
Our method matches it (55.9\%; paired McNemar test, $p{=}1.00$ on each engine) at 4.6$\times$ lower cost (\$0.69 \vs \$3.22 per task)
and with 2.3$\times$ (UE5) and 1.8$\times$ (Three.js) lower latency (\cref{sec:exp:efficiency}).
Per family, the two differ in both directions: ours is lower on UE5 spatial consistency (44.4\% \vs 63.0\%)
and higher on Three.js static physics (75.0\% \vs 57.1\%).

\begin{figure}[t]
  \centering
  \includegraphics[width=\linewidth]{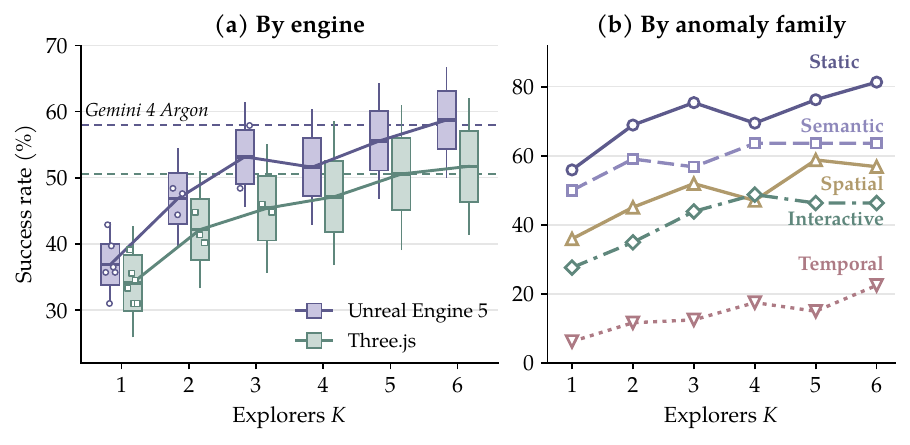}
  \caption{\textbf{Scaling the number of explorers $K$.} (a)~Success rate per engine, with single runs as dots. Boxes show $\pm$1
  standard error, whiskers 95\% bootstrap intervals over tasks, and dashed lines Gemini~4 Argon exploring alone. (b)~Per anomaly family.}
  \label{fig:scaling}
\end{figure}

\subsection{Scaling the Number of Explorers}
\label{sec:exp:scaling}

\cref{fig:scaling} varies the number of explorers $K$ with the review fixed.
For $K{<}6$ we review disjoint subsets of the six explorer runs and report their mean.
The success rate rises from 35.8\% at $K{=}1$ to 55.9\% at $K{=}6$, while the cost rises from \$0.19 to \$0.69 per task;
each added explorer costs about \$0.10, mostly for reviewing its trajectory.
Our method matches Gemini~4 Argon exploring alone at $K{=}6$ on UE5 and at $K{=}5$ on Three.js.
Every family benefits (\cref{fig:scaling}b): static physics gains 25.5 points and spatial consistency 21.0,
while temporal consistency rises from 6.2\% to 22.5\% but remains the hardest family, as in WorldAuditBench.

\begin{table}[t]
  \centering
  \small
  \setlength{\tabcolsep}{3.5pt}
  \renewcommand{\arraystretch}{1.12}
  \caption{\textbf{Ablations.} Success rate (\%) and cost (\$ per task). The frontier model is Gemini~4 Argon, the small model
  our fine-tuned Qwen3.5-4B. Rows reuse our six explorer runs unless the explorer changes, and single-explorer rows average six
  runs. 
  }
  \label{tab:ablation}
  \resizebox{\linewidth}{!}{
  \begin{tabular}{l|cc|c|c}
    \toprule
    \rowcolor{CadetBlue!15}
    \textbf{Variant} & \textbf{UE5} & \textbf{Three.js} & \textbf{Overall}$\uparrow$ & \textbf{Cost}$\downarrow$ \\
    \midrule
    \multicolumn{5}{l}{\textit{Division of the audit}} \\
    Frontier agent alone & 57.9\,{\scriptsize \blue{0.8}} & 50.6\,{\scriptsize \blue{1.1}} & 54.9\,{\scriptsize \blue{0.9}} & 3.22 \\
    \quad + our review of its trajectory & 56.3\,{\scriptsize \blue{2.4}} & 49.4\,{\scriptsize \blue{2.3}} & 53.5\,{\scriptsize \blue{2.3}} & 3.34 \\
    One small agent alone & 14.4\,{\scriptsize \blue{44.3}} & 17.6\,{\scriptsize \blue{34.1}} & 15.7\,{\scriptsize \blue{40.1}} & 0.03 \\
    Six small agents, reports pooled & 37.3\,{\scriptsize \blue{21.4}} & 39.1\,{\scriptsize \blue{12.6}} & 38.0\,{\scriptsize \blue{17.8}} & 0.18 \\
    \rowcolor{gray!10} Six small + Gemini 4 Argon review \textit{(ours)} & 58.7 & 51.7 & 55.9 & 0.69 \\
    \midrule
    \multicolumn{5}{l}{\textit{Explorer training (Qwen3.5-4B)}} \\
    Untrained, one explorer & \phantom{0}2.0\,{\scriptsize \blue{56.7}} & \phantom{0}6.3\,{\scriptsize \blue{45.4}} & \phantom{0}3.8\,{\scriptsize \blue{52.1}} & 0.02 \\
    Untrained, six + review & 28.6\,{\scriptsize \blue{30.2}} & 47.1\,{\scriptsize \blue{4.6}} & 36.2\,{\scriptsize \blue{19.7}} & 0.57 \\
    \rowcolor{gray!10} Teacher SFT, six + review \textit{(ours)} & 58.7 & 51.7 & 55.9 & 0.69 \\
    \midrule
    \multicolumn{5}{l}{\textit{Review (six SFT explorers)}} \\
    Single pass over all frames & 51.6\,{\scriptsize \blue{7.1}} & 46.0\,{\scriptsize \blue{5.7}} & 49.3\,{\scriptsize \blue{6.6}} & 0.45 \\
    Per explorer, then final (40 frames) & 54.8\,{\scriptsize \blue{4.0}} & 50.6\,{\scriptsize \blue{1.1}} & 53.1\,{\scriptsize \blue{2.8}} & 0.65 \\
    \rowcolor{gray!10} Per explorer, then final (96 frames) \textit{(ours)} & 58.7 & 51.7 & 55.9 & 0.69 \\
    \quad + full-res., high-effort first stage & 61.9\,{\scriptsize \red{3.2}} & -- & -- & 1.66 \\
    \bottomrule
  \end{tabular}
  }
\end{table}

\subsection{Ablations}
\label{sec:exp:ablation}

\cref{tab:ablation} changes one component at a time.

\paragraph{Division of the audit.}
The first group compares the designs of \cref{sec:method:division}.
A frontier agent exploring alone succeeds on 54.9\% of the tasks at \$3.22 each, and a single small agent on 15.7\%.
Six small agents together see far more anomalies, but pooling their own reports reaches only 38.0\%, because they recognize few of
the anomalies they see (\cref{sec:exp:analysis}).
Reviewing their trajectories with the frontier model raises the success rate to 55.9\% at \$0.69. The small agents provide the
coverage, and the frontier model the recognition.
The same review applied to the frontier agent's own trajectory gives 53.5\% at \$3.34 per task, 1.4 points below the agent's own
reports ($p{=}0.51$). A second look at a single strong trajectory does not help.

\paragraph{Explorer training.}
Without review, the untrained Qwen3.5-4B succeeds on 3.8\% of tasks and our explorer, fine-tuned on teacher trajectories, on 15.7\%.
With six explorers and our review, replacing our fine-tuned explorers with the untrained model lowers the overall success rate from 55.9\% to 36.2\%.
The drop is large on UE5 (58.7\% $\rightarrow$ 28.6\%, $p{<}10^{-8}$) and small on Three.js (51.7\% $\rightarrow$ 47.1\%, $p{=}0.48$),
because Three.js scenes are smaller (\cref{tab:difficulty}). There, 32.2\% of the anomalies are in view in the first frame and the
median one lies 8.7\,m from the start, so even an untrained explorer reaches 63.6\% of them. In UE5, no anomaly is in view at the start,
the median distance is 14.2\,m, and an untrained explorer reaches only 23.7\%, against 67.5\% after fine-tuning.

\paragraph{Review.}
A single pass over all frames of the six trajectories reaches 49.3\% at \$0.45 per task.
Reviewing each trajectory first and then deciding in a final review raises the success rate to 53.1\% with 40 frames in the final review
and to 55.9\% with 96 frames (UE5: 51.6\% $\rightarrow$ 58.7\%, $p{=}0.05$).
Reviewing each trajectory at full resolution with high effort adds 3.2 points on UE5 ($p{=}0.42$) at 2.3$\times$ the cost.

\begin{figure}[t]
  \centering
  \includegraphics[width=\linewidth]{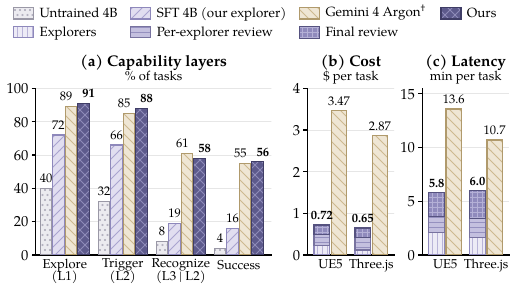}
  \caption{\textbf{Capability layers, cost, and latency.} (a)~Share of tasks in which the agent brings the anomaly into view (explore)
  and triggers it (trigger), share of triggered anomalies it reports correctly (recognize), and success rate. (b)~Cost per task at list
  prices. (c)~Mean model time per task with one explorer per H100 GPU.}
  \label{fig:layers}
\end{figure}

\begin{figure*}[t]
  \centering
  \includegraphics[width=\textwidth]{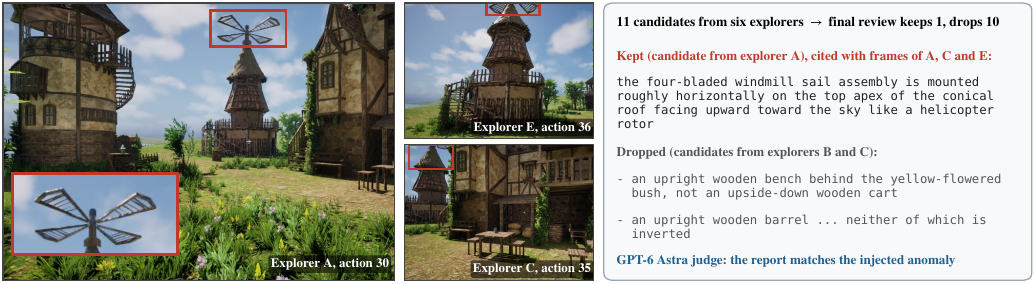}
  \caption{\textbf{Example from Medieval Village (UE5).} The windmill's sails are mounted horizontally on the roof (red boxes). No
  explorer reports it. The review of explorer A proposes it, and the final review confirms it with frames of explorers A, C, and E and
  drops ten others.}
  \label{fig:qual}
\end{figure*}

\subsection{Exploring, Triggering, and Recognizing}
\label{sec:exp:analysis}

We decompose auditing into three layers and measure each from the recorded trajectories (\cref{sec:app:layers}).
\emph{Explore}: the anomaly enters the view, \ie at least one frame places it within 10\,m and inside the field of view, following
WorldAuditBench's geometric coverage.
\emph{Trigger}: the agent also performs the behaviour the anomaly needs before it can show: approaching or touching the object for
collision anomalies, viewing it from different distances or angles for view-dependent anomalies, and looking away and back, or watching
for at least 10\,s, for temporal anomalies; anomalies that are visible without any action count as triggered once explored.
\emph{Recognize}: among triggered anomalies, the share that the agent reports correctly.

\cref{fig:layers} shows that fine-tuning mainly teaches the small model to explore and trigger: the explored share rises from 40.0\% to
72.3\% and the triggered share from 32.3\% to 65.7\%, approaching Gemini~4 Argon (88.7\% and 85.0\%).
Recognition improves less, from 7.7\% to 19.0\% of the triggered anomalies, far below Gemini~4 Argon's 61.3\%.
Our method combines the two strengths: six explorers explore 91.1\% and trigger 88.3\% of the anomalies, and the frontier review
recognizes 58.0\% of the triggered ones; it finds 50 of the 132 anomalies that no explorer reported itself.
Temporal anomalies remain the hardest: six explorers reach 87.5\% of them, but only 22.5\% are found.

\begin{table}[t]
  \centering
  \small
  \setlength{\tabcolsep}{3pt}
  \renewcommand{\arraystretch}{1.1}
  \caption{\textbf{Embodied question answering} on the 184 A-EQA questions of OpenEQA~\cite{majumdar2024openeqa}, without EQA
  training. LLM-Match accuracy (\%) per question category.
  Categories are attribute recognition (AR), functional reasoning (FR), object localization (OL), object recognition (OR), object
  state recognition (OSR), spatial understanding (SU), and world knowledge (WK).}
  \label{tab:eqa}
  \resizebox{\linewidth}{!}{
  \begin{tabular}{l|ccccccc|cc}
    \toprule
    \rowcolor{CadetBlue!15}
    \textbf{Method} & AR & FR & OL & OR & OSR & SU & WK & \textbf{All}$\uparrow$ & $\Delta$ \\
    \midrule
    \multicolumn{10}{l}{\textit{Answer from the question alone}} \\
    \logo{gemini}\model{Gemini 4 Argon} & 27.3 & 52.9 & 42.1 & 22.0 & 59.3 & 44.7 & 31.2 & 38.9 & \red{45.5} \\
    \midrule
    \multicolumn{10}{l}{\textit{Frontier model exploring alone}} \\
    \logo{openai}\model{GPT-6 Astra} & 70.5 & \second{73.5} & 80.7 & 79.0 & \best{95.4} & 64.5 & 77.7 & 78.0 & \red{6.4} \\
    \logo{claude}\model{Claude Opus 5} & \second{83.3} & 55.9 & 71.4 & 72.0 & 73.1 & 64.5 & 77.7 & 72.7 & \red{11.7} \\
    \logo{gemini}\model{Gemini 3.8 Flash} & 79.5 & \second{73.5} & 73.6 & 86.0 & 88.9 & \second{71.1} & 66.1 & 77.2 & \red{7.2} \\
    \logo{meta}\model{Muse Spark 1.3} & 72.0 & 66.2 & 69.3 & 68.0 & 75.9 & 56.6 & 61.6 & 67.8 & \red{16.6} \\
    \logo{gemini}\model{Gemini 4 Argon} & 81.8 & \best{76.5} & \second{85.7} & \best{92.0} & \second{92.6} & 69.7 & \best{86.6} & \best{84.5} & \blue{0.1} \\
    \midrule
    \rowcolor{gray!10} \textsc{Ours} & \best{84.8} & \second{73.5} & \best{87.1} & \second{87.0} & \second{92.6} & \best{73.7} & \second{83.9} & \second{84.4} & -- \\
    \bottomrule
  \end{tabular}
  }
\end{table}

\subsection{Efficiency}
\label{sec:exp:efficiency}

Panels (b) and (c) of \cref{fig:layers} break down where cost and time go.
At list prices our method costs \$0.72 per task on UE5 and \$0.65 on Three.js, 4.8$\times$ and 4.4$\times$ less than Gemini~4 Argon
exploring alone (\$3.47 and \$2.87).
Gemini~4 Argon re-reads its growing context at every decision, 1.6M input tokens per UE5 task over 26 model calls on average,
whereas our reviews read each trajectory once.
The six explorers account for only \$0.22 and \$0.11 at the OpenRouter price, and for nothing beyond GPU time when served locally.
Exploration also runs in parallel: with one explorer per H100 GPU, the slowest of the six explorers needs 2.1\,min (UE5) and 1.6\,min
(Three.js) of model time.
Including the review, our method takes 5.8 and 6.0\,min per task on average (medians 5.4 and 5.6\,min), whereas Gemini~4 Argon, which
decides every action in sequence, takes 13.6 and 10.7\,min (medians 9.5 and 7.7\,min). Ours is faster on 161 of the 213 tasks.

\subsection{Generalization to Embodied QA}
\label{sec:exp:eqa}

We test whether the same system, without any EQA training, can explore real-world scans to answer questions.
We use the 184-question A-EQA subset of OpenEQA~\cite{majumdar2024openeqa} and run every agent in Habitat~\cite{savva2019habitat} on
HM3D scenes~\cite{ramakrishnan2021habitat} with the same movement and viewing actions and the same budget of 40 actions.
Our explorers act as on WorldAuditBench, and the review answers the question instead of reporting anomalies.
We compare with five frontier models that explore alone with the same tools.
GPT-6 Astra runs in Codex and Claude Opus~5 in Claude Code, and the other models call the tools through their APIs.
Answering from the question alone reaches only 38.9\% (\cref{tab:eqa}), so the questions require exploring the scene.
Our method reaches 84.4\%, on par with Gemini~4 Argon exploring alone (84.5\%) and 6.4 to 16.6 points above GPT-6 Astra,
Gemini~3.8 Flash, Claude Opus~5, and Muse Spark~1.3.
It ranks first or second in each of the seven question categories.

\subsection{Qualitative Example}
\label{sec:exp:qual}

\cref{fig:qual} shows how the review combines evidence across explorers.
Several explorers capture the windmill, yet none of them reports its misplaced sails.
The review of explorer A's trajectory proposes the anomaly, and the final review confirms it with frames that explorers A, C, and E took
from different positions, while rejecting the ten other candidates, most of them ordinary objects that a per-explorer review mistook
for misplaced or upside-down ones.

\section{Conclusion}
\label{sec:conclusion}

We studied how to divide the audit of a 3D world between small and frontier models so that the auditor is both accurate and cheap. A 4B VLM fine-tuned end to end on verified trajectories of a teacher guided by Bayesian search learns bug seeking navigation and triggering, and six such explorers bring more glitches into view and trigger more of them than a frontier model exploring alone. Recognition stays with a frontier model, which reviews what the explorers record and never acts in the world. On WorldAuditBench, this auditor succeeds on 55.9\% of the tasks, 13.6 points above the strongest auditor reported, and matches Gemini~4 Argon exploring alone at 4.6$\times$ lower cost and about half the latency. Without further training, it also answers embodied questions about real scans as well as Gemini~4 Argon exploring alone. Recognition is now the main gap, especially for temporal anomalies, and better recognition in small models would let them take on more of the audit. A cheap and accurate auditor can also serve as a reward model for training agents that build interactive 3D worlds and games, a promising direction that we leave to future research.

\section{Acknowledgments}
\label{sec:ack}

The UCSB team acknowledges support from the National Science Foundation (NSF) under Grants IIS-2338252, IIS-2302730, and BCS-2619240.

{
    \small
    \bibliographystyle{ieeenat_fullname}
    \bibliography{main}
}

\clearpage
\setcounter{page}{1}
\maketitlesupplementary
\noindent Citations refer to the reference list of the main paper.
\appendix

\section{Training Worlds}
\label{sec:app:worlds}

\begin{figure*}[t]
  \centering
  \includegraphics[width=\textwidth]{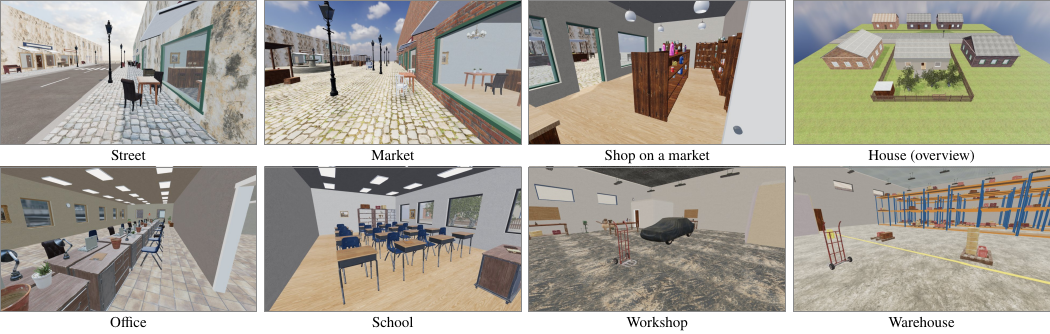}
  \caption{\textbf{Three.js training worlds}, one view of each place type, rendered from eye height by the plausibility gate.}
  \label{fig:worlds}
\end{figure*}

\paragraph{Three.js generator.}
Every Three.js world passes through six automatic stages. First, a language model (Gemini~3.8 Flash) fills a JSON schema that contains only the options the generator supports. For streets and markets these are the six shop types, the number of storeys, the length of the street, and the widths and textures of sidewalks and roads, for houses the four rooms, which include a living room and a kitchen, and for every place its footprint and a sky that suits its setting. Numbers snap to supported values, and invalid fields fall back to seeded random choices. Second, a seeded generator builds the ground, the building shells, and the backdrop buildings, and then furnishes the rooms one by one. Sidewalks, alleys, yards, and squares count as outdoor rooms bounded by curbs, facades, or fences, so the same anchors apply outdoors. Third, the world is assembled as a Three.js scene in which every placed object is a named group, which is what an anomaly injector targets. The physical check with repair, the plausibility gate, and the acceptance follow below. \cref{tab:places} summarizes the place types, and \cref{fig:worlds} shows one view of each.

\begin{table*}[t]
  \centering
  \footnotesize
  \renewcommand{\arraystretch}{1.1}
  \caption{\textbf{Place types of the Three.js training worlds.} A seeded generator builds each world from its plan and samples the sizes from the given ranges.}
  \label{tab:places}
  \begin{tabular}{l|p{3.4cm}|p{11.4cm}}
    \toprule
    \rowcolor{CadetBlue!15}
    \textbf{Place} & \textbf{Size} & \textbf{Content} \\
    \midrule
    Street & 36--48\,m long, road 7--9\,m & Two rows of terraced buildings with six open shops (sales room, storage, display windows) and closed units, sidewalks with lamps, benches, and cafe terraces, a back alley with dumpsters and pallets \\
    Market & a street with a 14--18\,m square & The shops of a street, a fountain, rows of stalls with produce and crates, planters \\
    House & 11--14 $\times$ 9--11\,m & A hall and four rooms including a living room and a kitchen, a fenced front yard, a backyard with patio and shed, the road and neighboring houses \\
    Office & 22--26 $\times$ 15--17\,m & Lobby, private office, meeting room, corridor, open office with desk islands, storage, kitchenette, a plaza in front \\
    School & 26--30 $\times$ 16.5\,m & Entrance hall, four classrooms, library, teachers' room, corridor, a fenced schoolyard \\
    Workshop & hall 15--18 $\times$ 12--14\,m & Workbenches with tools and pegboards, machines, a covered car, office, store room, a fenced yard \\
    Warehouse & hall 24--28 $\times$ 20--24\,m & Racks of pallets with 3\,m aisles, a staging area, dock doors, an office booth, a loading yard \\
    \bottomrule
  \end{tabular}
\end{table*}

\paragraph{Assets.}
The worlds draw on a library of 520 models, 81 textures, and 23 sky maps from Poly Haven~\cite{polyhaven}, all released under CC0. The 55 training worlds use 244 of the models, 54 textures, and 7 skies, together with 20 kinds of procedural props built from textured boxes, such as shelving, counters, fences, market stalls, and signs. A vision language model inspects every model from four rendered sides and records its front direction, and models that are not upright, broken, not a single object, or implausibly sized are dropped (27 of 469). Rays cast downward on an $11{\times}11$ grid measure the support levels of 135 models, such as table tops and shelf boards, with the height, extent, and free height above each level, and two further ray tests mark whether a level is open from the front and from above.

\paragraph{Placement rules.}
Objects enter a world only through anchors, so nothing stands loose in the open unless it would in reality. An object placed against a wall takes a free slot 3\,cm from the wall, away from doors, corners, other objects, and windows, unless it is lower than the sill. Other anchors place objects in corners, under the ceiling with the lowest point of a lamp at least 2.05\,m above the floor, on walls at typical heights such as 1.6--1.75\,m for pictures, on facades, facing a wall or a point, or on the measured levels of another object. Recipes for 17 room types and 14 shop types combine these anchors into functional groups. In a living room, for example, the sofa stands against the wall with the most free length, the TV stand faces it from the opposite wall, and the coffee table stands 0.45\,m in front of the sofa on a rug. Pupil desks stand in columns 1.4\,m apart and face the board with the first row 3.1\,m from it, shop counters stand 1.0\,m in front of the back shelving, and warehouse racks stand back to back with 3\,m aisles. Small objects enter only on levels that are large and open enough, keep 4\,cm apart, and are placed 2\,mm above the surface. Footprints are rotated rectangles that may not overlap, with margins that depend on the item, and a zone 1.4\,m deep on both sides of every open doorway stays free. After each table group, shelf row, or desk island, a walker of radius 0.25\,m must still reach every door of the room and at least 85\% of its free floor, otherwise the group is undone. A world holds 487 objects on average, from 136 to 1,275, and 53\% of them rest on another object.

\paragraph{Checks and acceptance.}
The physical check runs in headless Chromium. Per object, $5{\times}5$ downward rays find floating objects with a gap of more than 4\,cm (12\,cm for plants and rocks), sunken objects with a median gap below $-8$\,cm, and objects without support. Bounding boxes shrunk by 1\,cm followed by an exact triangle test with three-mesh-bvh find interpenetrating objects and objects inside walls. A flood fill from the spawn with a capsule of radius 0.25\,m on a 0.1\,m grid measures the reachable share of every room. Flagged objects are deleted together with what rests on them, and the check repeats up to twice. The plausibility gate renders every room from eye height and asks a vision language model to list implausible objects, such as floating, misplaced, misoriented, or misscaled ones, each with a point in the image that a raycast maps to an object. Every major issue is checked again on a close-up from another angle, and confirmed objects are deleted. The model also rates the realism of each view from 1 to 5. A world is accepted when its mean realism is at least 3 with no view below 2, the final physical check finds nothing, every room is at least 75\% reachable, and at most $\max(3, 3\%)$ of the objects are unreachable. In the 55 training worlds the repair removed 22 objects and the gate 153, and the final realism averages 3.38.

\paragraph{Rendering and episodes.}
Worlds are rendered with three.js r169 in headless Chromium at $960{\times}540$ pixels with a vertical field of view of $72^\circ$ from an eye height of 1.7\,m. A sky map provides the background and image-based lighting, together with a hemisphere light and one sun that casts shadows. Collision uses one static bounding volume hierarchy of all meshes except glass and trees, and the player is a capsule of radius 0.25\,m. A \texttt{move} covers 0.3--4\,m at a walking speed between 2.35 and 5.71\,m/s, \texttt{turn} and \texttt{look} rotate by up to $180^\circ$ and $90^\circ$, and \texttt{wait} lasts 0.5--5\,s. The engine captures a film frame every 0.5\,s of simulated time and pauses the world while the model decides. In 70\% of the episodes the audit is confined to a square of $\pm 12$\,m around the anomaly, and the rest cover the whole world. The start lies 6--15\,m from the anomaly, 0.8--7\,m for some anomalies that need contact, or at the entrance of the world. Targets are reachable objects outside the context zones of a world, such as the buildings across the road. They are either free standing, with a footprint of 0.3--2.6\,m and a height of 0.3--2.4\,m, or small objects on a surface, and each type adds its own conditions, such as a matching object in the same room for a shrunk object or a clear straight path on both sides for missing collision.

\paragraph{UE5 environments.}
The UE5 worlds come from two commercial environments, an office building and a street district rendered by day and by night. Each of the three maps is split into regions of walkable area around a start pose, and the training data uses 65 regions. A scan records the walkable area of each region and, for every object, the positions from which it is visible, and an object can become a target when it is visible from at least three positions. Before a task is used, a test inside the packaged game confirms that its anomaly works. The test measures the offset, scale, rotation, material, or shadow flag of a static anomaly, sweeps a capsule of the player's size through collision anomalies with and without the anomaly, looks at view and distance anomalies from probe points near and far, and for temporal anomalies looks at the object, turns or walks away, and returns. It also confirms that the ground lies within 10\,cm below the start pose.

\section{Anomaly Types}
\label{sec:app:types}

\cref{tab:types} lists the 44 anomaly types of the training data with their mechanism, the behavior that triggers them, and the number of accepted trajectories in each engine. Types implemented the same way in both engines share a row. Each task of the training data contains one injected anomaly, except 24 UE5 tasks that contain none and whose correct report is empty.

\begin{table*}[p]
  \centering
  \scriptsize
  \setlength{\tabcolsep}{4pt}
  \renewcommand{\arraystretch}{1.08}
  \caption{\textbf{Anomaly types} of the training data. Engine: T for Three.js, U for UE5. Records: accepted teacher trajectories (Three.js / UE5). Ranges give the parameters of the accepted tasks.}
  \label{tab:types}
  \begin{tabular}{l|c|p{7.3cm}|p{3.3cm}|r}
    \toprule
    \rowcolor{CadetBlue!15}
    \textbf{Anomaly} & \textbf{Engine} & \textbf{Mechanism} & \textbf{Trigger} & \textbf{Records} \\
    \midrule
    \multicolumn{5}{l}{\textit{Geometry}} \\
    Floating object & T, U & Raised by 0.21--1.40\,m, with what rests on it & none & 109 / 131 \\
    Floating small object & T & A small item 0.21--0.44\,m above its surface & none & 70 / -- \\
    Sunken object & T, U & 0.26--0.54\,m or 35--55\% of its height into the floor & none & 48 / 14 \\
    Sunken small object & T & 35--65\% of its height into its surface & none & 12 / -- \\
    Tilted object & T, U & Tilted by $15$--$40^\circ$ about its base & none & 67 / 32 \\
    Intersecting copy & T, U & A second copy 0.10--0.98\,m away that intersects it & none & 82 / 24 \\
    Object in a wall & T & Pushed 0.25--0.5\,m into the nearest wall & none & 38 / -- \\
    Giant object & T, U & Scaled by 1.8--4.0 about its base & none & 57 / 86 \\
    Shrunk object & T, U & Scaled by 0.35--0.6, next to an object of the same kind & none & 37 / 32 \\
    Squashed or stretched object & T & One axis scaled by 0.4--0.65 or 1.5--2.0 & none & 25 / -- \\
    \midrule
    \multicolumn{5}{l}{\textit{Semantic}} \\
    Upside-down object & T, U & Rotated by $180^\circ$ about a horizontal axis & none & 120 / 76 \\
    Object facing a wall & U & Turned by $180^\circ$ about the vertical axis & none & -- / 11 \\
    \midrule
    \multicolumn{5}{l}{\textit{Appearance}} \\
    Missing texture & T & Unlit plain magenta & none & 32 / -- \\
    Flat color & T & One unlit plain color & none & 32 / -- \\
    Placeholder texture & T & Checkerboard pattern repeated 2--8 times & none & 40 / -- \\
    Glowing object & T & Neon emission of strength 1.5--3.0 & none & 23 / -- \\
    Mirror-like object & T & Polished metal with roughness 0.03--0.09 & none & 22 / -- \\
    Semi-transparent object & T & Opacity 0.26--0.44 & none & 14 / -- \\
    Wrong material & U & All material slots replaced by another surface of the scene & none & -- / 56 \\
    Drawn through occluders & T & Drawn without depth test, so it shows through what stands in front & view it behind an occluder & 159 / -- \\
    Missing shadow & T, U & Casts no shadow, placed in sunlight & none & 8 / 7 \\
    Displaced shadow & T, U & The shadow comes from an invisible copy 0.8--2.0\,m away & none & 17 / 7 \\
    Distance-dependent shape & T & Drawn stretched or squashed beyond 3.2--3.6\,m & approach from afar & 4 / -- \\
    Distance-dependent proxy & T & Replaced by a plain box beyond 2.6--4.0\,m & approach from afar & 27 / -- \\
    Distance-dependent size & U & Drawn scaled by 0.5 or 1.7 beyond 5.0--7.9\,m & approach from afar & -- / 81 \\
    Invisible beyond a distance & T, U & Not drawn beyond 4.0--7.5\,m, collision kept & approach from afar & 18 / 36 \\
    Invisible when near & T & Not drawn within 2.5--3.4\,m & approach & 9 / -- \\
    Invisible from one side & T & Not drawn when seen from one of four directions & view it from several sides & 5 / -- \\
    Invisible when looked at & U & Hidden within $12.6$--$21.4^\circ$ of the view center & look at it, then slightly away & -- / 41 \\
    \midrule
    \multicolumn{5}{l}{\textit{Collision}} \\
    Missing collision & T, U & The player passes through the object & walk into it & 66 / 16 \\
    Invisible wall & T, U & An invisible box 0.8--4.1\,m wide blocks a passage or doorway & walk into it & 128 / 50 \\
    Violent contact reaction & T & Topples or flies off and bounces 2--5 times when touched & touch it & 40 / -- \\
    \midrule
    \multicolumn{5}{l}{\textit{Temporal}} \\
    Vanishes when unobserved & T, U & Disappears for good once the player looks away after seeing it & see it, look away, look back & 82 / 68 \\
    Appears when unobserved & T & An object appears in an empty place after the player looks away & see the place, look away, look back & 46 / -- \\
    Vanishes after a visit & T & Disappears after the player comes within 2.5--3.4\,m and leaves beyond 4.4--5.5\,m & approach, leave, return & 9 / -- \\
    Vanishes after a revisit & U & Disappears after the player leaves beyond 8--10\,m and returns & leave and return & -- / 4 \\
    Moves when unobserved & T & Moves by 1.2--2.2\,m whenever it is out of view & look away and back & 36 / -- \\
    Moves after a revisit & U & Shifted by 0.8--1.2\,m after the player leaves and returns & leave and return & -- / 4 \\
    Turns when unobserved & T & Turns by $90^\circ$, seats by $180^\circ$ & look away and back & 26 / -- \\
    Resizes when unobserved & T & Scaled by 0.46--0.65 or 1.35--1.51 & look away and back & 33 / -- \\
    Changes color when unobserved & T & Takes one of seven colors, sometimes without texture & look away and back & 61 / -- \\
    Changes material after a revisit & U & Takes another material after the player leaves and returns & leave and return & -- / 1 \\
    Day and night switch & T & The whole scene switches between day and night every 2.5--6\,s & wait and watch & 102 / -- \\
    Jumping sun & T & The sun and the shadows jump between two times of day every 2.6--6\,s & wait and watch & 94 / -- \\
    \bottomrule
  \end{tabular}
\end{table*}

\section{Trajectory Verification}
\label{sec:app:verify}

\begin{figure*}[t]
  \centering
  \includegraphics[width=\textwidth]{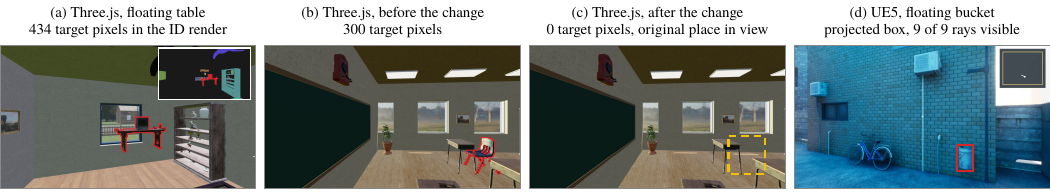}
  \caption{\textbf{Ground truth used for acceptance.} (a) The object-ID render of a Three.js frame (inset) gives the visible pixels of the target, here a floating table with the objects on it (outlined). (b, c) For a chair that vanishes after the teacher looks away, an earlier frame shows the chair and a later frame shows its former place empty (dashed box). (d) In UE5, the engine records the projected box of the target and the visibility of nine sample points, here for a floating bucket.}
  \label{fig:verify}
\end{figure*}

\paragraph{Ground truth per frame.}
After every action, the Three.js engine renders an object-ID image of $240{\times}135$ pixels with the same camera, in which every asset has its own color, and counts the pixels of the target together with the objects that rest on it or copy it (\cref{fig:verify}). It also projects the original bounding box of an object that changes over time, so that a later frame can show its former place, and logs every event of the anomaly, such as the player passing through the target, a contact with an invisible wall, the moment an object vanishes, or a switch of the light. The UE5 engine records for every frame the projected bounding box of the target and nine sample points, the eight corners of the box shrunk by 0.7 and its center. A sample point counts as visible when a ray from the camera reaches it or the target itself, or ends within 15\,cm of it. The record also holds the distance to the target and whether the target is currently drawn, and an event log covers the same kinds of events as in Three.js.

\paragraph{Acceptance in Three.js.}
A trajectory is accepted when its report matches the injected anomalies one to one. A finding matches an anomaly when its family is that of the anomaly, its description names the target with a word of its label or a synonym, and its cited frames show the anomaly. For a static anomaly, a cited frame has to show at least 60 target pixels in the ID render, about 0.2\% of the image, or be an evidence frame that the engine marked while the target was clearly in view (at least 130 pixels within 7\,m). For an object that vanishes, an earlier cited frame has to show at least 40 target pixels and a later one, after the change, none while its former place is in view, with the reverse for an object that appears and at least 40 pixels in both frames for an object that moves, turns, resizes, or changes color. For collision, view, and distance anomalies, a cited frame has to be among the frames that the engine marked when the anomaly showed, and for changes of daylight the cited frames have to cover two lighting states. A trajectory is discarded when any visible turn uses the vocabulary of the privileged context or copies a phrase of six or more words from it.

\paragraph{Acceptance in UE5.}
A trajectory is accepted when the teacher calls \texttt{done} within the budget, the engine confirms the injection and records the trigger, such as passing through an object without collision or being blocked by an invisible wall, and the report matches the injected anomalies one to one, which in a task without an anomaly means an empty report. The cited frames have to show a static anomaly in a clear view, in which the target is drawn, at least two of the nine sample points are visible, and the projected box covers at least 600 square pixels with a shorter side of at least 8 pixels. A temporal anomaly needs a clear view before the change and a view after it, a culling anomaly needs a frame in which the target is in view but not drawn, and a collision anomaly needs a frame within 2.5\,m of the object or the invisible wall. A language model (Gemini~3.8 Flash) then reads the finding together with the description of the injected anomaly and checks that it names the target, describes the same defect, and blames no other object.

\paragraph{Unintended glitches.}
Strong-model audits of the Three.js worlds sometimes report objects other than the injected targets. A vision language model judges each of them on a close-up in which the object is outlined in red, with the question: \emph{``An automatic auditor looked at frames of a 3D simulation of a real place and reported this problem about the object outlined in red. Is the reported problem really there? The renderer is a game engine, so ignore rendering quality. Views can mislead. Answer confirmed=true only if the problem is clearly visible on the outlined object and would look wrong in a real place.''} Confirmed objects are moved below the ground together with what rests on them before the training trajectories are generated, which removed 123 objects in 45 of the 55 worlds.

\paragraph{Acceptance rates.}
In the Three.js worlds, 31.0\% of the 3,665 trajectories of one generation run pass, from 98\% for changes of daylight and 61\% for appearance anomalies of materials to 11\% for missing collision and 9\% for shadow anomalies. In the UE5 worlds, about 39\% of the trajectories pass.

\section{Teacher Context}
\label{sec:app:teacher}

The privileged context consists of a search procedure, the same for every type, and the card of each injected type. Both are appended to the teacher's system prompt and never enter the training data. The search procedure reads as follows.

\begin{quote}
\scriptsize
\textsc{Exploration method} (private to you; follow it for the whole episode). 1. Survey. At the start, and whenever you enter a new room or area, turn around in place before walking anywhere, so you have seen the whole area. 2. Candidates. From what you have seen, decide which objects and places could show the assigned bug type (see the bug-type guide below). These are priorities, not limits: once the likely ones are checked and clean, check the other objects too. 3. Order. Go next to what is most likely to show the defect, easiest to check and closest. Skip what you have already seen clearly from up close. 4. Inspect and test. Get close enough to see the relevant part clearly (about 2--4\,m for furniture, closer for small items), aim at it (look down or up as needed) and use a second angle if unsure. \texttt{inspect} with a region zooms into a small detail; the film frames of an action show what happened during it. Do the test the bug type needs (see the guide). 5. Evidence. As soon as a frame shows something wrong, call \texttt{flag\_bug} with the frame refs that show it (for a change or a movement test, the before frame and the after frame). Confirm it with \texttt{update\_bug} after a closer look or a repeated test; retract a suspicion that turns out to be normal. 6. Memory. Keep track of what you have already checked; do not re-check without a reason (looking back for a change is a reason). 7. Next area. When everything likely in the current area is checked, go through the nearest unexplored doorway or opening and start again at step 1. 8. Finish with \texttt{done} once the candidates in the reachable areas have been checked and every report has been verified; there is no need to use all actions. Writing: every reply starts with one or two short sentences in your own words about what you see and what you will do next. Never mention instructions, rules, a method, a guide, an advisor, suggestions, candidate lists, rankings, scores, numbered steps or a checklist.
\end{quote}

The type card for invisible walls reads as follows.

\begin{quote}
\scriptsize
\textsc{Where it tends to show up} (priorities, not limits). Open walkable space that should be passable: doorways, corridors, gaps between furniture, open floor, roads, paths. Example: on a clear road, a forward move stops after a few centimetres. \textsc{How to check.} Walk through doorways and across open areas; when a move stops much shorter than requested and nothing visible is in the way, repeat the same move once to rule out a small obstacle. \textsc{Frames to cite.} The view before the blocked move and the view after it (barely changed), with the moved distance. \textsc{Not this bug.} Being stopped by a visible obstacle, a step, a door frame edge or furniture.
\end{quote}

\section{Results by engine}
\label{sec:app:engine}

\cref{tab:main_by_engine} reports the success rate by engine and anomaly family, in the format of WorldAuditBench's per-engine table.
Some cells rest on very few tasks; for example, Three.js has only two semantic-consistency tasks.

\begin{table*}[t]
  \centering
  \small
  \setlength{\tabcolsep}{3.4pt}
  \renewcommand{\arraystretch}{1.15}
  \caption{\textbf{Results by engine and anomaly family} (\cref{tab:main} pools the families over both engines). Success rate (\%), overall success rate (SR) on all
  213 tasks, and inference cost (\$ per task, list prices). SP: static physics; IP: interactive physics; SpC: spatial consistency;
  TC: temporal consistency; SeC: semantic consistency (tasks per family: 31/22/27/26/20 in Unreal Engine~5, 28/19/24/14/2 in Three.js).
  Rows without $^\dagger$ are copied from WorldAuditBench~\cite{jiang2026worldauditbench}; $^\dagger$ marks an auditor we run under the
  same protocol. VLA--VLM auditors explore with Open-P2P 1.2B and analyze the recorded trajectory with the listed VLM. Our method uses
  six Qwen3.5-4B explorers fine-tuned with SFT and Gemini~4 Argon as the reviewer; run-to-run variation is reported in
  \cref{tab:variance}. $\Delta$: our success rate minus the row's (points) and our cost minus the row's (\$)
  (\textcolor{RedOrange}{orange}: ours better, \textcolor{BlueGreen}{teal}: ours worse). \textbf{Bold} and \underline{underline}: best
  and second best per column. Human auditors reach 83.4\% overall~\cite{jiang2026worldauditbench}.}
  \label{tab:main_by_engine}
  \resizebox{\textwidth}{!}{
  \begin{tabular}{cl|ccccc|ccccc|cc|cc}
    \toprule
    \rowcolor{CadetBlue!15}
    \cellcolor{white} & \cellcolor{white}
    & \multicolumn{5}{c|}{\textbf{Unreal Engine 5}}
    & \multicolumn{5}{c|}{\textbf{Three.js}}
    & \multicolumn{2}{c|}{\textbf{Overall}}
    & \multicolumn{2}{c}{\textbf{Cost} (\$/task)} \\
    \cmidrule(lr){3-7} \cmidrule(lr){8-12} \cmidrule(lr){13-14} \cmidrule(lr){15-16}
    \cellcolor{white} & \cellcolor{white} \multirow{-2.25}{*}{\textbf{Method}}
    & SP & IP & SpC & TC & SeC & SP & IP & SpC & TC & SeC & SR$\uparrow$ & $\Delta$ & \$$\downarrow$ & $\Delta$ \\
    \midrule
    \multirow{6}{*}{\makecell{\textsc{VLM}\\\textsc{Auditors}}}
    & \logo{openai}\model{GPT-6 Astra} & 71.0 & 31.8 & \second{44.4} & 15.4 & \best{75.0} & 46.4 & 26.3 & 45.8 & \phantom{0}7.1 & \phantom{0}0.0 & 42.3 & \red{13.6} & 2.955 & \gooddown{2.26} \\
    & \logo{claude}\model{Claude Opus 5} & 25.8 & 13.6 & 29.6 & \second{19.2} & 55.0 & 46.4 & 26.3 & 29.2 & \phantom{0}0.0 & \phantom{0}0.0 & 28.2 & \red{27.7} & 2.445 & \gooddown{1.75} \\
    & \logo{gemini}\model{Gemini 3.8 Flash} & 51.6 & 18.2 & 22.2 & \second{19.2} & 55.0 & 46.4 & 26.3 & 37.5 & \phantom{0}0.0 & \phantom{0}0.0 & 32.4 & \red{23.5} & 1.382 & \gooddown{0.69} \\
    & \logo{meta}\model{Muse Spark 1.3} & 16.1 & \phantom{0}9.1 & \phantom{0}0.0 & \phantom{0}7.7 & 25.0 & 28.6 & 15.8 & 29.2 & \phantom{0}0.0 & \phantom{0}0.0 & 15.0 & \red{40.9} & 0.791 & \gooddown{0.10} \\
    & \logo{qwen}\model{Qwen 3.8 Flash} & \phantom{0}9.7 & \phantom{0}9.1 & \phantom{0}0.0 & \phantom{0}3.8 & 25.0 & 14.3 & \phantom{0}5.3 & \phantom{0}8.3 & \phantom{0}0.0 & \phantom{0}0.0 & \phantom{0}8.5 & \red{47.4} & 0.077 & \badup{0.62} \\
    & \logo{gemini}\model{Gemini 4 Argon}$^\dagger$ & \best{90.3} & \second{45.5} & \best{63.0} & 15.4 & \second{70.0} & \second{57.1} & 42.1 & \second{66.7} & \best{28.6} & \phantom{0}0.0 & \second{54.9} & \red{1.0} & 3.225 & \gooddown{2.53} \\
    \midrule
    \multirow{5}{*}{\makecell{\textsc{VLA--VLM}\\\textsc{Auditors}}}
    & \logo{openai}\model{GPT-6 Astra} & \phantom{0}9.7 & \phantom{0}0.0 & 11.1 & \phantom{0}0.0 & \phantom{0}5.0 & 32.1 & 36.8 & 41.7 & \phantom{0}0.0 & \best{100.0} & 16.4 & \red{39.5} & 0.281 & \badup{0.41} \\
    & \logo{claude}\model{Claude Opus 5} & \phantom{0}6.5 & \phantom{0}0.0 & \phantom{0}0.0 & \phantom{0}0.0 & \phantom{0}5.0 & 39.3 & \second{47.4} & 25.0 & \phantom{0}7.1 & \best{100.0} & 15.0 & \red{40.9} & 0.727 & \gooddown{0.03} \\
    & \logo{gemini}\model{Gemini 3.8 Flash} & 16.1 & \phantom{0}0.0 & \phantom{0}0.0 & \phantom{0}0.0 & 10.0 & 25.0 & \best{52.6} & 37.5 & \second{14.3} & \best{100.0} & 17.4 & \red{38.5} & 0.296 & \badup{0.40} \\
    & \logo{meta}\model{Muse Spark 1.3} & \phantom{0}3.2 & \phantom{0}0.0 & \phantom{0}0.0 & \phantom{0}0.0 & 10.0 & \phantom{0}7.1 & 21.1 & 20.8 & \phantom{0}0.0 & \phantom{0}0.0 & \phantom{0}6.6 & \red{49.3} & 0.112 & \badup{0.58} \\
    & \logo{qwen}\model{Qwen 3.8 Flash} & \phantom{0}9.7 & 13.6 & \phantom{0}0.0 & \phantom{0}0.0 & \phantom{0}5.0 & 28.6 & 31.6 & 25.0 & \phantom{0}0.0 & \phantom{0}0.0 & 12.7 & \red{43.2} & 0.039 & \badup{0.65} \\
    \midrule
    \rowcolor{gray!10}
    \multicolumn{2}{c|}{\textsc{Ours}} & \second{87.1} & \best{59.1} & \second{44.4} & \best{30.8} & \second{70.0} & \best{75.0} & 31.6 & \best{70.8} & \phantom{0}7.1 & \phantom{0}0.0 & \best{55.9} & -- & 0.694 & -- \\
    \bottomrule
  \end{tabular}
  }
\end{table*}

\section{Run-to-run variation}
\label{sec:app:variance}

WorldAuditBench reports one run per auditor, so \cref{tab:main} compares single runs.
\cref{tab:variance} reports how much our method varies across runs.
Across six independent explorer runs, a single explorer with our review reaches 35.8\,$\pm$\,3.0\% overall.
Repeating the review of our six explorer trajectories twice more from scratch gives 56.3\,$\pm$\,0.8\% overall
(runs: 55.9\%, 57.3\%, and 55.9\%), so the run in \cref{tab:main} is representative; most of the variation comes from Three.js
(54.0\,$\pm$\,2.3\%, \vs 57.9\,$\pm$\,0.8\% on UE5).

\begin{table}[t]
  \centering
  \small
  \setlength{\tabcolsep}{4.5pt}
  \renewcommand{\arraystretch}{1.15}
  \caption{\textbf{Run-to-run variation of our method.} Success rate (\%), mean\,$\pm$\,standard deviation over $n$ runs.
  Top: independent explorer runs, each reviewed once. Bottom: the review of our six explorer trajectories repeated from scratch.
  WorldAuditBench reports one run per auditor, so \cref{tab:main} compares single runs.}
  \label{tab:variance}
  \resizebox{\linewidth}{!}{
  \begin{tabular}{l|c|cc|c}
    \toprule
    \rowcolor{CadetBlue!15}
    \textbf{Setting} & $n$ & \textbf{UE5} & \textbf{Three.js} & \textbf{Overall} \\
    \midrule
    \multicolumn{5}{l}{\textit{Different explorer runs}} \\
    Explorer only (one explorer, no review) & 6 & 14.4\std{0.93} & 17.6\std{4.28} & 15.7\std{1.67} \\
    \textsc{Ours} ($K{=}1$) & 6 & 36.9\std{4.04} & 34.1\std{3.06} & 35.8\std{2.98} \\
    \midrule
    \multicolumn{5}{l}{\textit{Same six explorer runs, review repeated}} \\
    \rowcolor{gray!10} \textsc{Ours} ($K{=}6$) & 3 & 57.9\std{0.79} & 54.0\std{2.30} & 56.3\std{0.81} \\
    \bottomrule
  \end{tabular}
  }
\end{table}

\section{Reach analysis}
\label{sec:app:reach}

Anomaly positions come from the benchmark: the tagged target actor of each Unreal Engine~5 task and the injected object of each
Three.js task. The environment records the position and view direction of every delivered frame; crops of earlier frames are excluded.
As a check of the coordinate conventions, the frames cited by correct reports face the anomaly (within $\pm$45$^\circ$ horizontally)
in 62.5\% (UE5) and 69.0\% (Three.js) of the cases; the rest are context views.
\cref{tab:reach} reports reach under distance limits of 7, 10, and 16\,m.

\begin{table}[t]
  \centering
  \small
  \setlength{\tabcolsep}{5pt}
  \renewcommand{\arraystretch}{1.15}
  \caption{\textbf{Reach under different distance limits} (\% of the 213 tasks). A task is reached if at least one frame places the
  anomaly within the distance limit and within $\pm$41.4$^\circ$ horizontally and $\pm$32$^\circ$ vertically of the view direction.}
  \label{tab:reach}
  \begin{tabular}{c|cc|c}
    \toprule
    \rowcolor{CadetBlue!15}
    \textbf{Limit} & \textbf{One explorer} & \textbf{Six explorers} & \textbf{Gemini 4 Argon}$^\dagger$ \\
    \midrule
    7\,m & 59.8 & 84.0 & 80.8 \\
    \rowcolor{gray!10} 10\,m & 72.3 & 91.1 & 88.7 \\
    16\,m & 87.9 & 97.7 & 94.4 \\
    \bottomrule
  \end{tabular}
\end{table}

\cref{tab:difficulty} explains why the choice of explorer matters more on UE5 than on Three.js (\cref{sec:exp:ablation}):
Three.js anomalies are closer to the start, so even an untrained explorer often reaches them.

\begin{table}[t]
  \centering
  \small
  \setlength{\tabcolsep}{5pt}
  \renewcommand{\arraystretch}{1.15}
  \caption{\textbf{Exploration difficulty by engine.} Three.js anomalies are closer to the start, so even an untrained explorer often
  reaches them; UE5 anomalies are never in view at the start and need a trained explorer. Reach uses the 10\,m limit of
  \cref{sec:app:reach}.}
  \label{tab:difficulty}
  \begin{tabular}{l|cc}
    \toprule
    \rowcolor{CadetBlue!15}
    & \textbf{UE5} & \textbf{Three.js} \\
    \midrule
    Anomaly in view in the first frame (\%) & 0.0 & 32.2 \\
    Anomaly within 10\,m of the start (\%) & 18.3 & 60.9 \\
    Median distance from the start (m) & 14.2 & 8.7 \\
    \midrule
    Reached by one untrained explorer (\%) & 23.7 & 63.6 \\
    Reached by one SFT explorer (\%) & 67.5 & 79.3 \\
    Reached by six untrained explorers (\%) & 59.5 & 80.5 \\
    Reached by six SFT explorers (\%) & 88.9 & 94.3 \\
    \bottomrule
  \end{tabular}
\end{table}

\section{Capability layers}
\label{sec:app:layers}

\cref{sec:exp:analysis} measures three layers per task from the frames and actions of each episode.
\emph{Explore} uses the reach criterion of \cref{sec:app:reach} with a 10\,m limit.
\emph{Trigger} depends on the anomaly's subcategory in the WorldAuditBench taxonomy.
Anomalies that show without any action (floating, intersecting, mis-scaled, lighting, material, and semantic anomalies) count as
triggered once explored.
For missing collisions, the agent must come within 3\,m of the object; for unexpected collisions, within 3\,m or be blocked within 5\,m;
for abnormal motion after interaction, within 3\,m or interact within 5\,m. Several collision targets are annotated as a review point
rather than the object itself, so we use these generous radii.
For visibility and view- or distance-dependent anomalies, the agent must see the target (within 16\,m and in view) from distances
spanning at least 4\,m or from directions at least 45$^\circ$ apart.
For temporal anomalies, the agent must see the target, lose it from view for at least 3\,s of simulated time, and see it again, or see
it over at least 10\,s.
\emph{Recognize} is the share of triggered tasks whose report is judged correct.
As a check, the trigger rule holds for 87.5\% of the correct reports on temporal anomalies, 86.5\% on view-dependent ones, and 67.4\%
on collision ones, whose targets are the least precise.
\cref{tab:layers} reports the layers on the tasks whose anomaly needs a trigger.

\begin{table}[t]
  \centering
  \small
  \setlength{\tabcolsep}{4pt}
  \renewcommand{\arraystretch}{1.15}
  \caption{\textbf{Capability layers on the 119 tasks whose anomaly needs a trigger} (collision, view- or distance-dependent, and
  temporal anomalies). Explore and trigger: \% of tasks; recognize: \% of triggered tasks reported correctly; SR: success rate.}
  \label{tab:layers}
  \resizebox{\linewidth}{!}{
  \begin{tabular}{l|ccc|c}
    \toprule
    \rowcolor{CadetBlue!15}
    \textbf{Agent} & \textbf{Explore} & \textbf{Trigger} & \textbf{Recognize} & \textbf{SR} \\
    \midrule
    Untrained Qwen3.5-4B & 44.1 & 30.4 & \phantom{0}8.7 & \phantom{0}3.9 \\
    SFT Qwen3.5-4B (our explorer) & 71.1 & 59.4 & 14.0 & 12.9 \\
    \model{Gemini 4 Argon}$^\dagger$ & 87.4 & 80.7 & 46.9 & 39.5 \\
    \rowcolor{gray!10} \textsc{Ours} & 89.9 & 84.9 & 41.6 & 39.5 \\
    \bottomrule
  \end{tabular}
  }
\end{table}

\section{Judge agreement on A-EQA}
\label{sec:app:eqa}

We score the A-EQA answers of the six agents that explore (\cref{tab:eqa}) a second time with GPT-6 Astra as judge.
The two judges give the same LLM-Match mark to 91.3\% of the 1,104 answers, differ by more than one mark on a single answer, and
their marks correlate at $r{=}0.98$.
With GPT-6 Astra as judge, the accuracies are 83.4\% for ours, 83.7\% for Gemini~4 Argon, 78.9\% for GPT-6 Astra, 77.2\% for
Gemini~3.8 Flash, 71.9\% for Claude Opus~5, and 67.9\% for Muse Spark~1.3.
Every accuracy changes by less than one point, and the order of the agents is the same under both judges.

\section{Cost accounting}
\label{sec:app:cost}

All costs are list prices per task; judge calls are excluded for every method, and the costs of WorldAuditBench's auditors are copied
from its paper.
With this accounting, our harness reproduces the cost that WorldAuditBench reports for Gemini~3.8 Flash on Three.js (\$1.45 \vs
\$1.48 per task).
Gemini~4 Argon is priced at \$2 per million input tokens and \$10 per million output tokens, including thinking tokens, using the
token counts the API reports for each call: every model call of an episode when it explores alone, and the per-explorer and final
reviews in our method.
OpenRouter does not list Qwen3.5-4B, so we price our explorers at the list price of Qwen3.5-9B, the smallest Qwen3.5 model it lists
(\$0.025 per million input tokens and \$0.036 per million output tokens).
This overestimates the cost of a 4B model; at this price the six explorers cost \$0.22 per UE5 task and \$0.11 per Three.js task.

\end{document}